\documentclass[11pt, a4paper]{lumia}

\usepackage[sort&compress]{natbib}
\usepackage{xspace}

\theoremstyle{plain}

\newtheorem*{proposition*}{Proposition}

\theoremstyle{definition}

\theoremstyle{definition}

\def\eqref#1{equation~\ref{#1}}

\usepackage{graphicx}           
\usepackage{tikz}               
\usepackage[edges]{forest}      

\usepackage{url}                
\usepackage{xurl}               

\usepackage{array}              
\usepackage{longtable}          
\usepackage{multirow}           
\usepackage{makecell}           
\usepackage{ragged2e}           

\usepackage{mathtools}          
\usepackage{nicefrac}           

\usepackage{algorithm}          
\usepackage{algorithmicx}       
\usepackage{algpseudocode}      
\usepackage{listings}           

\usepackage{subcaption}         
\usepackage{wrapfig}            
\usepackage[export]{adjustbox}  

\usepackage[commandnameprefix=always]{changes}            
\usepackage{xspace}             
\usepackage[normalem]{ulem}     
\usepackage{CJKutf8}            

\usepackage[tikz]{bclogo}       
\usepackage[framemethod=tikz]{mdframed} 

\usepackage{lipsum}             
\usepackage{tocloft}            
\usepackage{afterpage}          
\usepackage{bbding}             
\usepackage{epigraph}           
\usepackage{minitoc}            
\usepackage{multicol}           
\usepackage{textgreek}          

\usepackage{lineno}
\usepackage{bm}
\usepackage{cuted}

\newcolumntype{P}[1]{>{\RaggedRight\arraybackslash}p{#1}}

\definecolor{uclablue}{RGB}{39, 116, 174}
\definecolor{bigaired}{RGB}{156, 0, 0}
\definecolor{myblue}{HTML}{598BE7}
\definecolor{mildblue}{RGB}{31,119,180}
\definecolor{sectionblue}{RGB}{70, 130, 180}
\definecolor{methodblue}{RGB}{0, 150, 136}
\definecolor{bgblue}{RGB}{245,243,253}
\definecolor{ttblue}{RGB}{91,194,224}
\definecolor{mygreen}{rgb}{0.64, 0.56, 0.88}
\definecolor{myyellow}{rgb}{0.68, 0.6, 0.1}
\definecolor{fancygreen}{rgb}{0.33, 0.68, 0.20}
\definecolor{salmon}{rgb}{0.94, 0.52, 0.49}
\definecolor{tablegreen}{rgb}{0.82, 0.94, 0.75}
\definecolor{tableblue}{rgb}{0.81, 0.90, 0.94}
\definecolor{tablered}{rgb}{0.97, 0.85, 0.85}
\definecolor{tableorange}{rgb}{0.96, 0.85, 0.81}
\definecolor{myorange}{rgb}{1.0, 0.49, 0.0}
\definecolor{tlgreen}{rgb}{0.33, 0.68, 0.20}
\definecolor{darkgreen}{RGB}{0,100,0}
\definecolor{darkred}{RGB}{200, 0, 0}
\definecolor{customyellow}{HTML}{FFFACD}
\definecolor{refinegreen}{RGB}{0, 128, 75}
\definecolor{scoregreen}{RGB}{34, 139, 34}
\definecolor{hidden-blue}{RGB}{194,232,247}
\definecolor{hidden-black}{RGB}{20,68,106}
\definecolor{yes}{HTML}{C6EFCE}
\definecolor{no}{HTML}{FFC7CE}
\definecolor{partial}{HTML}{FFEB9C}
\definecolor{external}{HTML}{D9E1F2}
\definecolor{hdr}{HTML}{F2F2F2}
\definecolor{GRPOrow}{gray}{0.96}
\definecolor{FlowRLrow}{RGB}{225,236,255}
\definecolor{FlowBlue}{RGB}{80,120,210}
\definecolor{GRPOGray}{gray}{0.35}

\hypersetup{
    colorlinks=true, 
    citecolor=uclablue, 
    linkcolor=bigaired,
    urlcolor=darkblue
}

\setlist[itemize]{leftmargin=20pt, noitemsep, topsep=0pt}

\NewDocumentCommand{\kaiyan}{mO{}}{\textcolor{purple}{\textsuperscript{\textit{kaiyan}}\textsf{\textbf{\small[#1]}}}}
\NewDocumentCommand{\yuxin}{mO{}}{\textcolor{cyan}{\textsuperscript{\textit{yuxin}}\textsf{\textbf{\small[#1]}}}}
\NewDocumentCommand{\bx}{mO{}}{\textcolor{green}{\textsuperscript{\textit{bx}}\textsf{\textbf{\small[#1]}}}}
\NewDocumentCommand{\at}{mO{}}{\textcolor{red}{\textsuperscript{\textit{AT}}\textsf{\textbf{\small[#1]}}}}
\NewDocumentCommand{\re}{mO{}}{\textcolor{blue}{\textsuperscript{\textit{RE}}\textsf{\textbf{\small[#1]}}}}
\NewDocumentCommand{\ybsun}{mO{}}{\textcolor{magenta}{\textsuperscript{\textit{youbang}}\textsf{\textbf{\small[#1]}}}}
\NewDocumentCommand{\runze}{mO{}}{\textcolor{orange}{\textsuperscript{\textit{runze}}\textsf{\textbf{\small[#1]}}}}
\NewDocumentCommand{\add}{mO{}}{\textcolor{darkgreen}{\textsuperscript{\textit{Maybe Consider Discuss}}\textsf{\textbf{[#1]}}}}

\newcommand{\cmark}{\textcolor{darkgreen}{\boldmath$\checkmark$}}
\newcommand{\xmark}{\textcolor{darkred}{\boldmath$\times$}}

\newenvironment{itemize*}%
 {\leftmargini=10pt\begin{itemize}%
  \setlength{\itemsep}{0pt}%
  \setlength{\parskip}{0pt}%
  }%
 {\end{itemize}}

\newenvironment{enumerate*}%
 {\begin{enumerate}%
  \setlength{\itemsep}{0pt}%
  \setlength{\parskip}{0pt}}%
 {\end{enumerate}}

\newcommand{\cellstatus}[1]{%
  \begingroup
  \StrTrim{#1}[\statusval]%
  \IfStrEq{\statusval}{Yes}{\cellcolor{yes}\cmark}{}%
  \IfStrEq{\statusval}{No}{\cellcolor{no}\xmark}{}%
  \IfBeginWith{\statusval}{Yes (}{\cellcolor{yes}\cmark~\textit{\statusval\unskip}}{}%
  \IfStrEq{\statusval}{Partial}{\cellcolor{partial}\textbf{Partial}}{}%
  \IfStrEq{\statusval}{External}{\cellcolor{external}\textbf{External}}{}%
  \endgroup
}

\newtcolorbox{myboxi}[1][]{
  breakable,
  title=#1,
  colback=red!5,
  colbacktitle=red!5,
  coltitle=black,
  fonttitle=\bfseries,
  bottomrule=0pt,
  toprule=0pt,
  leftrule=2pt,
  rightrule=2pt,
  titlerule=0pt,
  arc=0pt,
  outer arc=0pt,
  colframe=red,
}

\newtcolorbox{myboxnote}[1][]{
  breakable,
  title=#1,
  colback=orange!0,
  colbacktitle=orange!0,
  coltitle=black,
  fonttitle=\bfseries,
  bottomrule=0pt,
  toprule=0pt,
  leftrule=2pt,
  rightrule=2pt,
  titlerule=0pt,
  arc=0pt,
  outer arc=0pt,
  colframe=orange,
}

\newtcolorbox{myboxii}[1][]{
  breakable,
  freelance,
  title=#1,
  colback=white,
  colbacktitle=white,
  coltitle=black,
  fonttitle=\bfseries,
  bottomrule=0pt,
  boxrule=0pt,
  colframe=white,
  overlay unbroken and first={
  \draw[red!75!black,line width=3pt]
    ([xshift=5pt]frame.north west) -- 
    (frame.north west) -- 
    (frame.south west);
  \draw[red!75!black,line width=3pt]
    ([xshift=-5pt]frame.north east) -- 
    (frame.north east) -- 
    (frame.south east);
  },
  overlay unbroken app={
  \draw[red!75!black,line width=3pt,line cap=rect]
    (frame.south west) -- 
    ([xshift=5pt]frame.south west);
  \draw[red!75!black,line width=3pt,line cap=rect]
    (frame.south east) -- 
    ([xshift=-5pt]frame.south east);
  },
  overlay middle and last={
  \draw[red!75!black,line width=3pt]
    (frame.north west) -- 
    (frame.south west);
  \draw[red!75!black,line width=3pt]
    (frame.north east) -- 
    (frame.south east);
  },
  overlay last app={
  \draw[red!75!black,line width=3pt,line cap=rect]
    (frame.south west) --
    ([xshift=5pt]frame.south west);
  \draw[red!75!black,line width=3pt,line cap=rect]
    (frame.south east) --
    ([xshift=-5pt]frame.south east);
  },
}

\tcbset{
  takeawaysbox/.style={
    title=Takeaways,
    colback=lightblue!80,
    colframe=black,
    fonttitle=\bfseries\small,
    coltitle=white,
    colbacktitle=black,
    enhanced,
    attach boxed title to top left={xshift=2.5mm,yshift=-2.5mm},
    boxed title style={rounded corners, size=small, colframe=black, colback=black},
    width=\linewidth,
    arc=3.5mm
  }
}

\mdfdefinestyle{mystyle}{%
  rightline=true,
  innerleftmargin=10,
  innerrightmargin=10,
  outerlinewidth=3pt,
  topline=false,
  rightline=true,
  bottomline=false,
  skipabove=\topsep,
  skipbelow=\topsep
}

\tikzset{%
    every node/.style={font=\tiny},
    parent/.style =          {align=center,text width=2cm,rounded corners=3pt, line width=0.3mm, fill=gray!10,draw=gray!80},
    child/.style =           {align=center,text width=2.0cm,rounded corners=3pt, fill=blue!10,draw=blue!80,line width=0.3mm},
    grandchild/.style =      {align=center,text width=2cm,rounded corners=3pt},
    greatgrandchild/.style = {align=center,text width=1.5cm,rounded corners=3pt},
    greatgrandchild2/.style = {align=center,text width=1.5cm,rounded corners=3pt},    
    referenceblock/.style =  {align=center,text width=1.5cm,rounded corners=2pt},
    pretrain/.style =           {align=center,text width=2.0cm,rounded corners=3pt, fill=blue!10,draw=blue!80,line width=0.3mm},   
    pretrain_work/.style =           {align=center, text width=8.5cm,rounded corners=3pt, fill=blue!10,draw=blue!0,line width=0.3mm},  
    template/.style =           {align=center,text width=2.0cm,rounded corners=3pt, fill=red!10,draw=red!80,line width=0.3mm},   
    template_work/.style =           {align=center,text width=8.5cm,rounded corners=3pt, fill=red!10,draw=red!0,line width=0.3mm},    
    answer/.style =           {align=center,text width=2.0cm,rounded corners=3pt, fill= cyan!10,draw= cyan!80,line width=0.3mm},   
    answer_work/.style =           {align=center,text width=8.5cm,rounded corners=3pt, fill= cyan!10,draw= cyan!0,line width=0.3mm},      
    multiple/.style =           {align=center,text width=2.0cm,rounded corners=3pt, fill= orange!10,draw= orange!80,line width=0.3mm},   
    multiple_work/.style =           {align=center,text width=8.5cm,rounded corners=3pt, fill= orange!10,draw= orange!0,line width=0.3mm},        
    tuning/.style =           {align=center,text width=2.0cm,rounded corners=3pt, fill= magenta!10,draw= magenta!80,line width=0.3mm},   
    tuning_work/.style =           {align=center,text width=8.5cm,rounded corners=3pt, fill= magenta!10,draw= magenta!0,line width=0.3mm},          
}

\newcommand{\lstbg}[3][0pt]{{\fboxsep#1\colorbox{#2}{\strut #3}}}

\lstdefinelanguage{diff}{
  basicstyle=\ttfamily\small,
  morecomment=[f][\lstbg{red!20}]-,
  morecomment=[f][\lstbg{green!20}]+,
}

\lstdefinelanguage{diffpython}{
  language=diff,
  morekeywords={def, if, else, for, while, return, import, from, as, class, with, try, except, finally, raise, lambda, and, or, not, in, is, None, True, False},
  morecomment=[l]{\#},
  morestring=[b]",
  morestring=[b]',
}

\setheadertext{LUMIA Lab}

\correspondingemail{\emailicon \href{mailto:json.kai@sjtu.edu.cn}{json.kai@sjtu.edu.cn} \quad $^\ddagger$ Corresponding Author. }

\renewcommand{\today}{2026-06-25}

\title{Information-Aware KV Cache Compression for Long Reasoning}
\setheadertitle{Information-Aware KV Cache Compression for Long Reasoning}

\author{%
  \Authfont Jushi Kai$^{1,2}$, Zhuiri Xiao$^{3}$, Alexandra Birch$^{2\ddagger}$, Zhouhan Lin$^{1\ddagger}$\\
  $^1$ LUMIA Lab, School of Artificial Intelligence, Shanghai Jiao Tong University \\
  $^2$ School of Informatics, University of Edinburgh \\
  $^3$ Shanghai Jiao Tong University
}

\begin{document}

\begin{abstract}
Reasoning capability has advanced rapidly in large language models (LLMs), leading to an increasing size of key-value (KV) cache in both prefilling and decoding stages. Existing KV cache compression methods mainly rely on attention weights to estimate token importance. While attention effectively captures contextual relevance, it overlooks complementary information-theoretic signals related to predictive uncertainty and token informativeness. In this paper, we revisit token importance from a forward-looking perspective and introduce \textit{Forward Influence}, a metric that measures how compressed tokens affect future contexts. Our analysis reveals that tokens selected by attention scores mainly influence nearby contexts, whereas tokens associated with high predictive uncertainty exhibit substantially stronger influence on distant future contexts. Based on the observation, we propose \textbf{InfoKV}, an entropy-aware KV cache compression framework that incorporates information-theoretic signals. It combines token-level predictive uncertainty with layer-wise representation evolution and integrates the resulting entropy scores with attention scores during reasoning. Experiments on long-context reasoning benchmarks with Llama-3.1, Llama-3.2, and DeepSeek-R1 demonstrate that InfoKV consistently outperforms existing attention-based KV compression methods in both long prefilling and decoding scenarios.
\footnote{We will release our code for reproducibility later.} 

\end{abstract}

\maketitle



\section{Introduction}

Large language models (LLMs) have demonstrated remarkable capabilities in long-context understanding and reasoning \citep{deepseek-r1, openai-o1}. However, their deployment in long-sequence processing remains computationally expensive due to the quadratic growth of computing attention and the linear growth of key–value (KV) cache memory \citep{DMS, rpc}. This bottleneck is especially pronounced in long-form reasoning tasks, where thousands of tokens are handled as inputs or outputs for LLMs.

To address this issue, recent studies have explored KV cache compression techniques that selectively retain only a subset of past tokens. A common paradigm estimates token importance based on attention weights from a fixed observation window, e.g., the most recent tokens \citep{snapkv, pyramidkv, rpc}. Tokens receiving larger attention weights from recent contexts are regarded as important and preserved, while the remaining tokens are discarded. Such strategies have shown promising improvements in inference efficiency and memory reduction.

Despite their effectiveness, attention-based KV compression methods suffer from an inherent limitation: they rely on \textbf{short-term, backward-looking} signals. Specifically, importance is inferred from the extent to which recent tokens attend to past tokens, which primarily captures local dependencies. However, long-form reasoning could depend on information that may not be directly activated by recent contexts but remains crucial for future reasoning trajectories. This mismatch becomes especially problematic in long-decoding scenarios, where reasoning paths evolve dynamically over generation steps.


\begin{wrapfigure}{r}{0.65\textwidth}
    \centering

    \begin{subfigure}[t]{0.32\columnwidth}
        \centering
        \includegraphics[width=\linewidth]{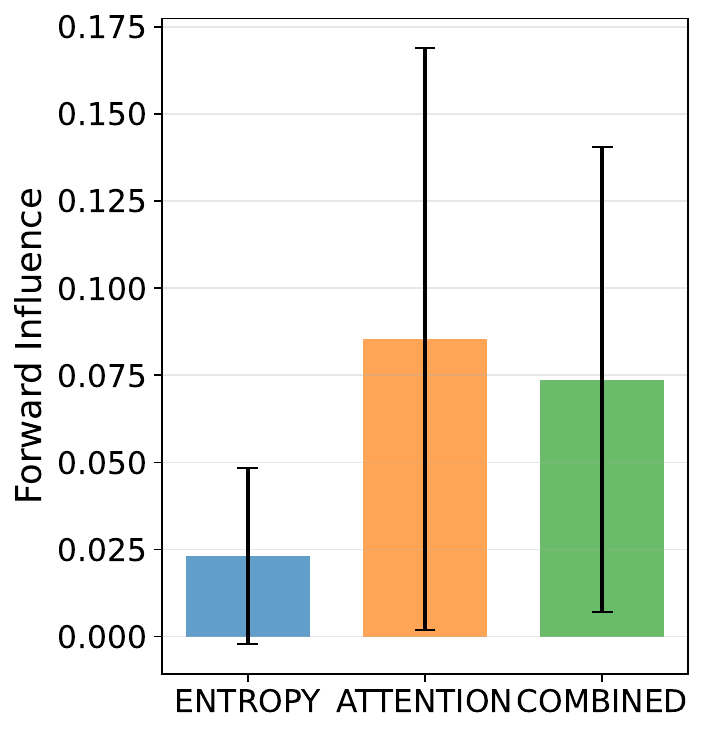}
        \caption{Short-range influence (128-token horizon).}
        \label{fig:short_range}
    \end{subfigure}
    \hfill
    \begin{subfigure}[t]{0.32\columnwidth}
        \centering
        \includegraphics[width=\linewidth]{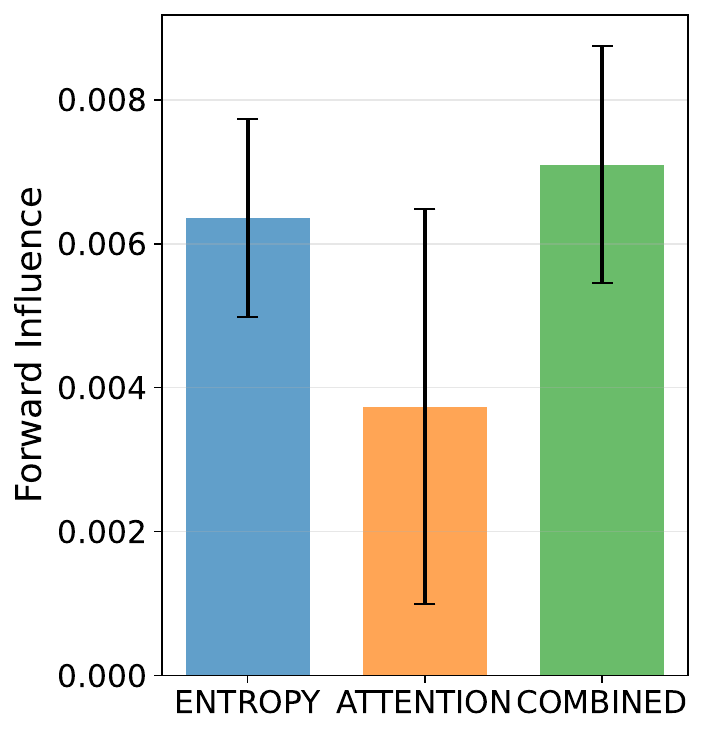}
        \caption{Long-range influence (14K-token horizon).}
        \label{fig:long_range}
    \end{subfigure}

    \caption{
    Comparison of short-range and long-range influence for top-1\% tokens scored by entropy, attention weight and their combination over 100 documents from Arxiv-Summarization \citep{arxiv}. Short-range influence captures immediate predictive effects, whereas long-range influence reflects persistent long-context impact.
    }
    
    \label{fig:short_long_influence}
\end{wrapfigure}


In this work, we show that effective KV cache compression should be guided by \textbf{forward-looking token utility}, namely, how much a token contributes to future generation steps rather than only its relevance to recent contexts. To investigate this phenomenon, we introduce \textit{Forward Influence}, which measures the divergence in future predictive distributions after removing a token from the KV cache. As shown in Figure~\ref{fig:short_long_influence}, while attention emphasizes tokens that are closely relevant to recent contexts, entropy measures the informativeness of tokens, and those of high entropy exhibit substantially stronger and more persistent influence on distant future contexts.

Motivated by this observation, we propose an entropy-aware KV cache compression framework \textbf{InfoKV} that incorporates information-theoretic signals into token selection. Since entropy reflects the uncertainty of the model when predicting tokens, it naturally captures tokens carrying richer semantic information. To further characterize token importance across layers, we combine entropy with the representational evolution between intermediate and final layers, which is orthogonal to the sequence dimension.

Extensive experiments on both long prefilling and long decoding benchmarks demonstrate that preserving informative tokens substantially improves reasoning performance. In long prefilling scenarios, InfoKV consistently outperforms existing attention-based KV cache compression methods on LongReason across different context lengths and cache budgets with Llama-3.1 and Llama-3.2. In long decoding scenarios, InfoKV further achieves substantial improvements on IFEval, AIME 2024, and LiveCodeBench with DeepSeek-R1, demonstrating its effectiveness for mathematical reasoning, instruction following, and code generation tasks.
\section{Related Work}

\paragraph{KV Cache Compression.}

A dominant line of KV cache compression research focuses on selectively evicting past tokens based on attention patterns. Recent methods such as SnapKV \citep{snapkv}, PyramidKV \citep{pyramidkv} and FastKV \citep{fastkv} propose heuristic pruning strategies that measure token importance by attention weights and discard less-attended tokens. Other works explore token merging to approximate the original attention of full cache \citep{cam,kvmerger,d2o}. Although these methods effectively reduce memory usage, they primarily rely on attention-based heuristics, which are inherently backward-looking and mainly take effect on long prefilling tasks with short answers. 

\paragraph{Compression for Long-decoding.}

The reasoning ability of LLMs has raised increasing attention in recent years \citep{deepseek-r1, openai-o1}. With long reasoning paths to be generated, decoding latency and KV cache growth become more critical than prefilling efficiency. To address this challenge, recent studies have extended KV cache compression from the prefilling stage to the decoding stage. RPC \citep{rpc} generalizes SnapKV \citep{snapkv} to online decoding by periodically compressing the KV cache throughout generation. Expected Attention \citep{expected-attention} further estimates the expected contribution of tokens to future attention. In addition, FreqKV \citep{freqkv} proposes an iterative frequency-domain compression framework that supports both prefilling and decoding compression, enabling efficient train-short-test-long capability.

\paragraph{Information Signals for Token Importance.}
Beyond attention-based heuristics, recent work has explored information-theoretic signals to characterize token importance from a more intrinsic perspective. Unlike attention weights, which depend on contextual interactions, the information that a token carries represents its native importance. Selective Context \citep{self-info} leverages self-information to quantify the informativeness of tokens and prune redundant content in LLM inputs. Building on uncertainty-based measures, \citet{sh2} propose SH2, which utilizes prediction uncertainty to identify informative tokens and adjust the output distribution for improved factuality. In the context of long-form reasoning, SeLaR \citep{SeLaR} incorporates entropy-aware contrastive regularization to encourage exploration by pushing representations away from over-confident predictions. In this paper, we introduce information signals to better reflect the token influence on future contexts and optimize KV cache compression for long-context reasoning.

\begin{figure*}[t]
    \centering
    \includegraphics[width=0.95\textwidth]{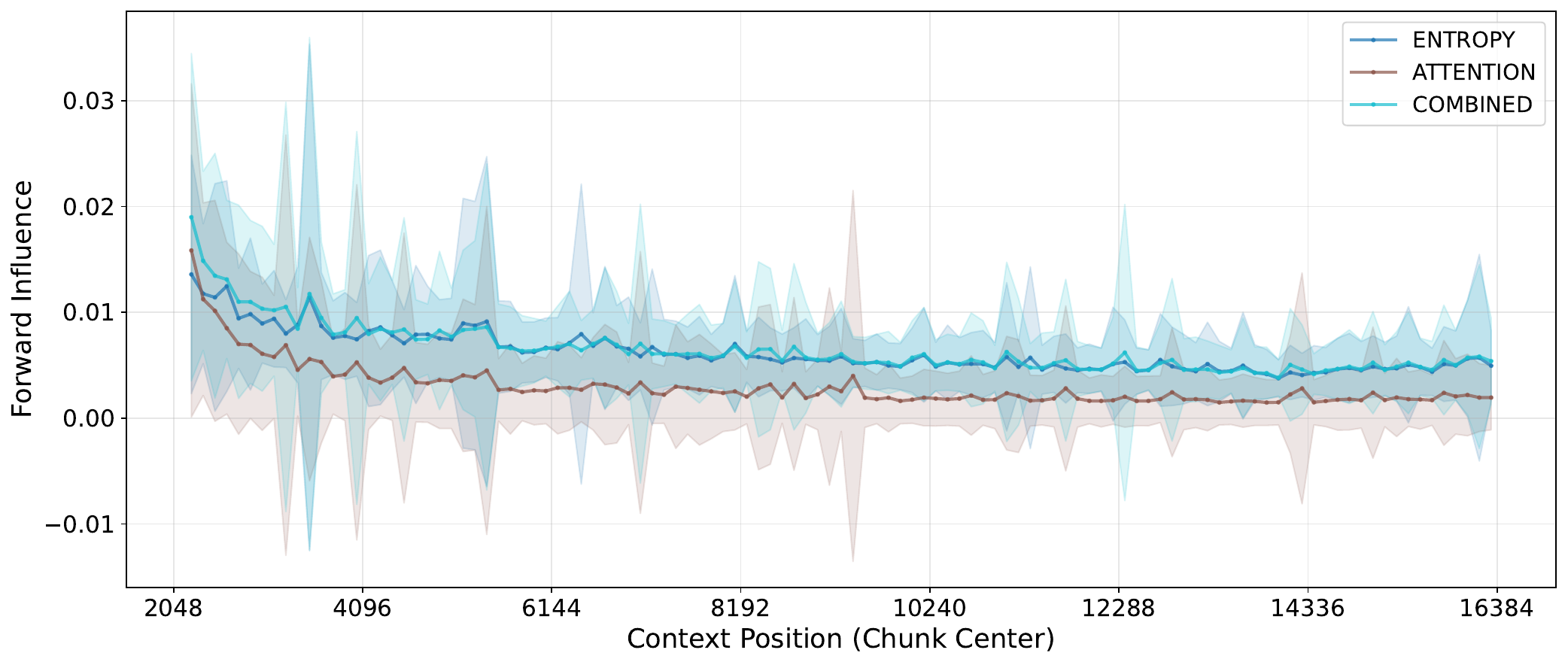}
    \caption{
    Forward influence of top-1\% tokens selected by different scoring strategies over long generation horizons on 100 documents from Arxiv-Summarization. The first 2048 tokens are compressed using different token importance scores, and the influence is measured over future chunks of 128 tokens. The combined score achieves a better balance between short-range and long-range influence.
    }
    \label{fig:influence_overview}
\end{figure*}

\section{Methodology}

\subsection{Revisiting Token Importance during Inference}

Existing KV cache compression methods predominantly estimate token importance according to attention scores computed from a recent observation window. Specifically, tokens receiving large attention weights from recent tokens are regarded as important and preserved in the KV cache. Although effective for maintaining short-range dependencies, such strategies implicitly assume that tokens important to recent contexts will remain important for future generation steps. However, during long-form reasoning and extended decoding, the relevance of tokens evolves continuously, and tokens with high recent attention scores may only contribute locally to nearby contexts while providing limited utility for future reasoning trajectories.

To better characterize the long-term utility of tokens, we revisit token importance from an information-theoretic perspective. Intuitively, tokens associated with high uncertainty carry more information for the language model and are therefore more likely to influence future contexts to be generated. As revealed in \citet{sh2}, these tokens are prone to be content words such as  such as adjectives, nouns, and conjugated verbs, which are more informative than function words like conjunctions, determiners and prepositions.

Given a sequence of tokens $\{x_0, x_1, \cdots, x_{n-1}\}$, the prediction probability of the next token $x_n$ by an autoregressive language model $\theta$ can be formalized as:
\begin{equation}
    \hat{p}(x_n) = p_\theta(x_n \mid x_{<n}).
\end{equation}

For the token $x_n$, we measure its uncertainty using the entropy of the predictive distribution:
\begin{equation}
H(x_n)
=
-
\sum_{x_n \in \mathcal{V}}
\hat{p}(x_n)
\log
\hat{p}(x_n),
\end{equation}
where $\mathcal{V}$ denotes the vocabulary space. A higher entropy indicates that the model is less confident when predicting the next token, implying that the corresponding context contains richer information. 

\begin{figure*}[t]
    \centering
    \includegraphics[width=1\textwidth]{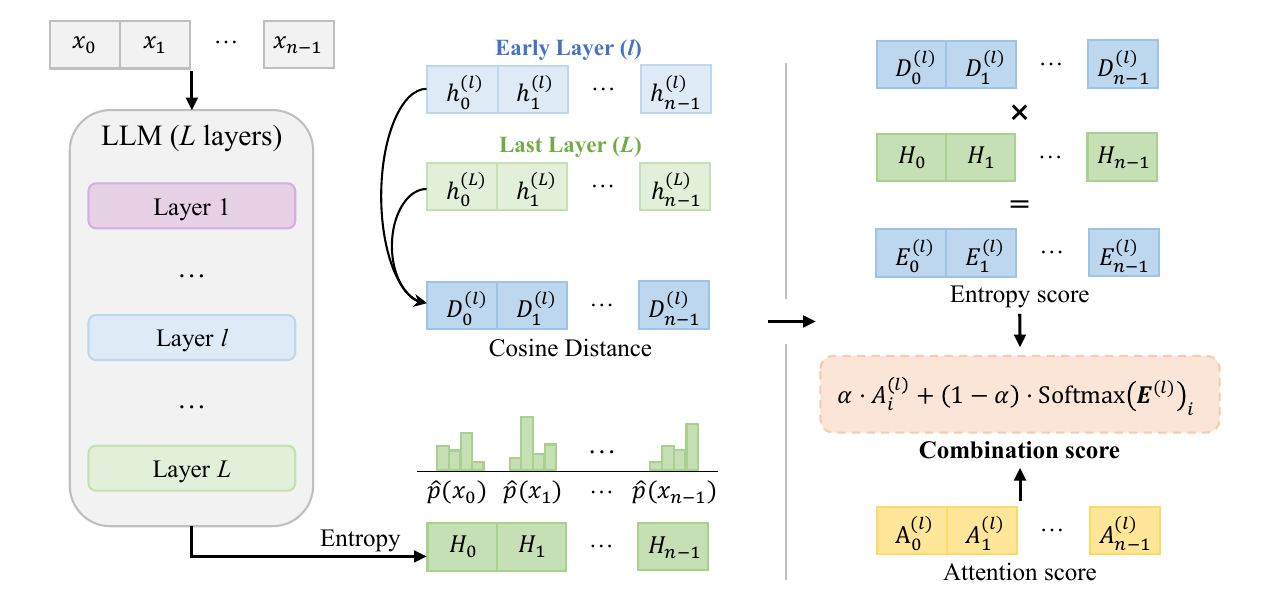}
    \caption{
    The overview of how to compute the importance score for KV cache compression in each layer. InfoKV combines predictive entropy, layer-wise representation evolution, and attention scores for token selection.
    }

    \label{fig:pipeline}
\end{figure*}

\subsection{Influence Estimation of Compressed KV Cache}

We conduct influence estimation with Llama-3.1-8B-Instruct \citep{llama3} to motivate our approach. We define the \textit{Forward Influence} of token $x_i$ in KV cache on a future context chunk $\{x_{l_c}, \cdots, x_{r_c}\}$, as the average divergence between the original prediction distribution and the prediction distribution obtained after removing $x_i$ from the KV cache:
\begin{equation}
\begin{aligned}
I_{l_c:r_c}(x_i)
=
\frac{1}{r_c-l_c+1}
&
\sum_{n=l_c}^{r_c}
\mathcal{D}_{\mathrm{KL}}
\Big(
p_\theta(x_n \mid x_{<n})\Big\|
p_\theta(x_n \mid x_{<n}\setminus\{x_i\})
\Big),
\end{aligned}
\end{equation}
where $p_\theta(x_n|x_{<n} \setminus \{x_i\})$ denotes the prediction probability of $x_n$ after removing $x_i$ from the KV cache. We use Kullback–Leibler (KL) divergence to quantify the difference between two predictive distributions. For simplicity, all layers share the same token choice for compression when estimating forward influence.

Based on this metric, we analyze the long-range contribution of tokens selected by different importance criteria, including attention scores, entropy, and their combinations. Specifically, we first rank tokens according to their averaged attention weights from a recent observation window $\{x_{l_o}, \cdots, x_{r_o}\}$, following prior KV cache compression methods:
\begin{equation}
A_i
=
\frac{1}{r_o-l_o+1}\sum_{t=l_o}^{r_o}\text{Attn}(q_t,k_i),
\end{equation}
where $\text{Attn}(q_t,k_i)$ is the attention weight from the token in the observation window to the token $x_i$ and extracted from the last layer.

We then compare them with tokens selected by entropy-based criteria. To combine attention weights and entropy, we use softmax to normalize the scale of entropy in the sequence dimension and add it to the attention score:
\begin{equation}
S_i
=
A_i+\text{Softmax}(\bm{H})_i.
\end{equation}

We use these three scoring strategies to compress the first 2048 tokens in a document and estimate their forward influence over a short future horizon and a long future horizon. Figure~\ref{fig:short_long_influence} demonstrates that the combination score balances short-range influence and long-range influence. 

Forward influence along the long sequence is presented in Figure~\ref{fig:influence_overview}. It reveals a clear distinction between attention-based and entropy-based importance estimation. Tokens with high attention scores mainly influence nearby future contexts, and their influence decays rapidly as the generation distance increases. In contrast, tokens with high entropy exhibit substantially stronger influence on distant future contexts, suggesting that entropy better captures information relevant to long-range reasoning and generation. By combining both signals, we can pick out tokens that are important for recent contexts as well as distant future contexts. Example visualizations are provided in Appendix~\ref{adp:case} to further illustrate the scoring difference between attention and entropy.

\subsection{KV Compression by Entropy}

We propose InfoKV to incorporate information signals into KV cache compression. As illustrated in Figure~\ref{fig:pipeline}, InfoKV integrates informativeness along the sequence dimension and semantic evolution across layers as the entropy score for each layer. The final importance score is the combination of the attention score and the entropy score.

To quantify token importance for each layer, we measure how much the hidden representation of a token evolves from an intermediate layer to the final layer. Specifically, for token $x_i$, we compute the cosine distance between the hidden states from the early layer $l$ and the final layer $L$:
\begin{equation}
D_i^{(l)}
=
1-
\cos
\left(
h_i^{(l)},
h_i^{(L)}
\right),
\end{equation}
where $h_i^{(l)}$ and $h_i^{(L)}$ denote the hidden representations of token $x_i$ at layer $l$ and the final layer $L$, respectively.

Due to the residual connections within Transformer architectures \citep{attention}, hidden representations evolve progressively across layers. If the representation at an early layer already closely aligns with the final-layer representation, the token has largely converged semantically and may contain limited additional information for future decoding. In contrast, tokens exhibiting larger representation shifts across layers tend to carry more unresolved semantic information and remain influential during subsequent generation.

We therefore combine the representation distance with the entropy computed from the final layer to estimate the entropy score for each token:
\begin{equation}
E_i^{(l)}
=
D_i^{(l)}
\cdot
H_i.
\end{equation}

Since the predictive probability mass of LLMs is typically concentrated on a small subset of highly probable tokens, these tokens dominate the model's decision-making process. Consequently, uncertainty estimated over the entire vocabulary can be heavily affected by numerous low-probability tokens that contribute little to generation behavior. Following prior work \citep{SeLaR}, we employ \textit{Top-$k$ Restricted Entropy} by using only the top-$k$ most probable tokens in the predictive distribution, which provides a more stable and informative estimation of uncertainty.

A bias $\tau$ will be added to $D_i^{(l)}$ so that the entropy score of the final layer will not be 0. Integrating token-level informativeness with layer-wise representation evolution, the entropy score jointly captures predictive uncertainty and the degree of representation transformation throughout layers. 

For each layer, we compute the token importance score by combining the attention score and the entropy score:
\begin{equation}
\label{eq:balanced_score}
S_i^{(l)}
=
\alpha \cdot A_i^{(l)}
+
(1-\alpha) \cdot \text{Softmax}(\bm{E}^{(l)})_i,
\end{equation}
where $A_i^{(l)}$ denotes the attention score of token $x_i$ at layer $l$, and $\mathbf{E}^{(l)}$ represents the entropy scores of all tokens in layer $l$. Given the importance scores $S^{(l)}$, we retain the top-ranked tokens in KV cache for each layer.

\newcommand{\rot}[1]{\multicolumn{1}{c}{\rotatebox{20}{#1}}}
\begin{table*}[!t]
\caption{Accuracy (\%) comparison on LongReason across different cache rates and prefill lengths (16k -- 64k). We mark the best scores in bold and underline the second-best scores in the table.}
\label{tab:LongReasoning}
\begin{center}
\setlength{\tabcolsep}{4pt} 
\begin{tabular}
{l@{\hspace{1pt}}c@{}c@{}c c@{}c c@{}c c@{}c}
\toprule
\multirow{3}{*}{\bf Rate} & \multirow{3}{*}{\bf Method} & \multicolumn{2}{c}{\bf 16k} & \multicolumn{2}{c}{\bf 32k} & \multicolumn{2}{c}{\bf 64k} & \multicolumn{2}{c}{\bf Ave.}\\
\cmidrule(lr){3-4} \cmidrule(lr){5-6} \cmidrule(lr){7-8} \cmidrule(lr){9-10}
& & \rot{w. CoT} & \rot{w/o. CoT} & \rot{w. CoT} & \rot{w/o. CoT} & \rot{w. CoT} & \rot{w/o. CoT} & \rot{w. CoT} & \rot{w/o. CoT} \\

\midrule
\rowcolor{gray!20}
\multicolumn{10}{c}{\bf \textit{Llama-3.1-8B-Instruct}} \\
100\% & Full 
& 55.67 & 52.08 
& 53.90 & 49.75 
& 53.02 & 48.61 
& 54.20 & 50.15 \\
\midrule

\multirow{4}{*}{40\%} 
& SnapKV 
& 53.15 & \underline{50.13} 
& \underline{51.13} & 45.72 
& \underline{48.99} & 45.59 
& 51.09 & \underline{47.15} \\

& PyramidKV 
& 53.67 & 47.36 
& 51.01 & \underline{46.35} 
& 47.61 & 45.04 
& 50.76 & 46.25 \\

& Expected 
& \underline{54.16} & 48.36 
& 50.50 & \underline{46.35} 
& 48.74 & \underline{45.72} 
& \underline{51.13} & 46.81 \\

& \bf InfoKV 
& \textbf{55.32} & \textbf{51.80} 
& \textbf{52.39} & \textbf{48.61} 
& \textbf{49.87} & \textbf{46.22} 
& \textbf{52.53} & \textbf{48.88} \\

\midrule

\multirow{4}{*}{20\%} 
& SnapKV 
& \textbf{52.39} & 47.23 
& 47.98 & 45.59 
& \underline{47.73} & 44.28 
& \underline{49.37} & 45.70 \\

& PyramidKV 
& 50.88 & \textbf{49.12} 
& \underline{48.36} & 44.84 
& 46.47 & 43.53 
& 48.57 & \underline{45.83} \\

& Expected 
& 50.88 & 45.72 
& 47.74 & \underline{45.97} 
& 47.48 & \underline{45.34} 
& 48.70 & 45.68 \\

& \bf InfoKV
& \underline{51.77} & \underline{48.74} 
& \textbf{49.50} & \textbf{46.22} 
& \textbf{48.11} & \textbf{45.72} 
& \textbf{49.79} & \textbf{46.89} \\

\midrule
\rowcolor{gray!20}
\multicolumn{10}{c}{\bf \textit{Llama-3.2-3B-Instruct}} \\

100\% & Full 
& 48.23 & 45.59 
& 46.47 & 44.96
& 44.96 & 42.70 
& 46.55 & 44.42 \\
\midrule

\multirow{4}{*}{40\%} 
& SnapKV 
& 45.34 & \underline{42.95} 
& 42.81 & \bf42.57 
& 40.93 & 39.42 
& 43.03 & 41.65
\\

& PyramidKV 
& 44.71 & \bf43.20 
& 43.07 & \underline{42.44}
& 40.81 & \underline{41.06}
& 42.86 & \bf 42.23
\\

& Expected 
& 45.09 & 42.82 
& 43.07 & 42.07 
& 41.18 & 39.29 
& \underline{43.11} & 41.39
\\

& \bf InfoKV
& \bf 46.47 & {42.44}
& \bf 43.82 & 41.69 
& \bf 41.56 &  \bf{41.31}
& \bf 43.95 & \underline{41.81}
\\



\midrule

\multirow{4}{*}{20\%} 
& SnapKV 
& 43.19 &  \bf 41.06
& 40.30 &  39.92
& 38.53 &  37.28
& 40.67 &  39.42 \\

& PyramidKV 
& \underline{43.42} &  40.18
& 40.42 &  39.67
& 38.41 &  \bf{38.28}
& 40.75 &  39.38 \\

& Expected 
& 42.44 &  \underline{40.55}
& \underline{40.93} &  \underline{40.05}
& \underline{38.79} &  37.66
& 40.72 &  39.42 \\

& \bf InfoKV
& \bf 43.83 &  \underline{40.55}
& \bf 41.69 &  \bf 40.43
& \bf 39.17 &  \underline{38.04}
& \bf 41.56 &  \bf 39.67\\ 

\bottomrule
\end{tabular}
\end{center}
\end{table*}

\begin{figure*}[t]
    \centering
    \includegraphics[width=0.94\textwidth]{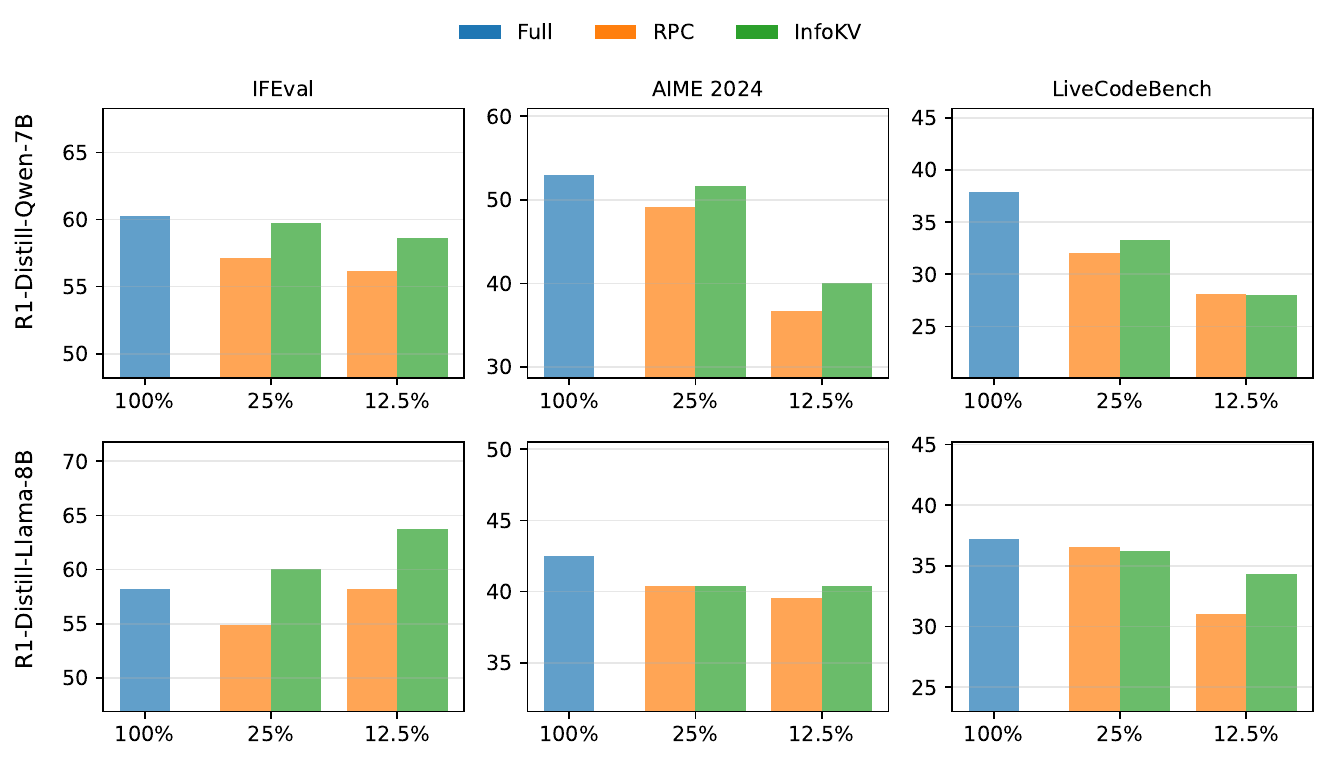}
    \caption{
    Performance on the three categories of long decoding benchmarks.
    }

    \label{fig:long_decoding}
\end{figure*}

\section{Experiments}

\subsection{Setup}

We assess InfoKV on both long prefilling and decoding scenarios. For long prefilling, we evaluate Llama-3.1-8B-Instruct and Llama-3.2-3B-Instruct  \citep{llama3} on the long-context reasoning benchmark LongReason \citep{longreason}. Models will process the entire input prompt in parallel and compress KV cache of the prompt for the following decoding stage.

As for long decoding, we employ InfoKV on reasoning models DeepSeek-R1-Distill-Qwen-7B and DeepSeek-R1-Distill-Llama-8B \citep{deepseek-r1}. Models are evaluated on IFEval \citep{ifeavl}, American Invitational Mathematics Examination (AIME) 2024 \citep{aime24}, and LiveCodeBench \citep{livecodebench}. Models will compress KV cache for the generated tokens periodically during the decoding phase.

\subsection{Long Prefilling}

We evaluate InfoKV on the long-context reasoning benchmark LongReason, which expands original reasoning tasks into long-context inputs containing extensive supporting evidence and distractor information. Thereby, it stresses the ability of KV cache compression methods to preserve reasoning- critical information under limited KV cache budgets. We compare InfoKV against three representative attention-based methods: SnapKV~\citep{snapkv}, which uses attention scores from a recent observation window; PyramidKV~\citep{pyramidkv}, which further introduces layer-wise budget allocation; and Expected Attention~\citep{expected-attention}, which estimates the expected contribution of tokens to future attention distributions. Experiments are conducted on Llama-3.1-8B-Instruct and Llama-3.2-3B-Instruct under different cache retaining ratios. Models are evaluated in chain-of-thought (w. CoT) and direct-answer (w/o. CoT) settings. For fair comparison, all baselines are implemented following the official implementations and share the same evaluation configurations. More details can be referred to in Appendix~\ref{apd:prefill_exp_setting}.

Experimental results across multiple context lengths are reported in Table~\ref{tab:LongReasoning}. Overall, InfoKV obtains SOTA (state-of-the-art) or highly competitive performance across most settings, demonstrating the effectiveness of incorporating entropy-aware information signals into KV cache compression. InfoKV consistently outperforms all attention-based baselines under both 40\% and 20\% cache budgets on Llama-3.1-8B-Instruct. The advantage becomes more evident as the sequence length increases. It suggests that entropy-aware token selection can better retain globally informative tokens that remain useful throughout long reasoning trajectories, whereas recent-attention heuristics tend to emphasize short-range dependencies and may discard information important for future reasoning steps.

\subsection{Long Decoding}

For long decoding, we consider IFEval for instruction following, AIME 2024 for mathematical reasoning, and LiveCodeBench for coding evaluation. Models are required to generate reasoning steps and derive final answers with a maximum output length of 32768 tokens. Following RPC~\citep{rpc}, which periodically compresses KV cache in the decoding stage based on attention weights, we sample 1 completion for IFEval, 8 completions per instance for AIME 2024, and 4 completions for LiveCodeBench to compute pass@1 scores. KV cache compression is triggered every 1024 tokens during decoding. Settings of hyperparameters are summarized in Appendix~\ref{apd:decode_exp_setting}.

Performance on  DeepSeek-R1-Distill-Qwen-7B and DeepSeek-R1-Distill-Llama-8B is shown in Figure~\ref{fig:long_decoding}. InfoKV achieves better performance on the three task categories compared to RPC. Notably, on IFEval, InfoKV with retaining ratios of 25\% and 12.5\% even surpasses the full cache of R1-Distill-Llama-8B. It suggests that long reasoning trajectories contain substantial redundancy, and retaining all historical tokens may introduce distracting or less informative contexts during generation. By selectively compressing tokens associated with high predictive certainty and lower information content, InfoKV enables the model to focus more effectively on informative reasoning contexts and improves generation quality.

\begin{figure}[t]
    \centering
    \includegraphics[width=0.48\textwidth]{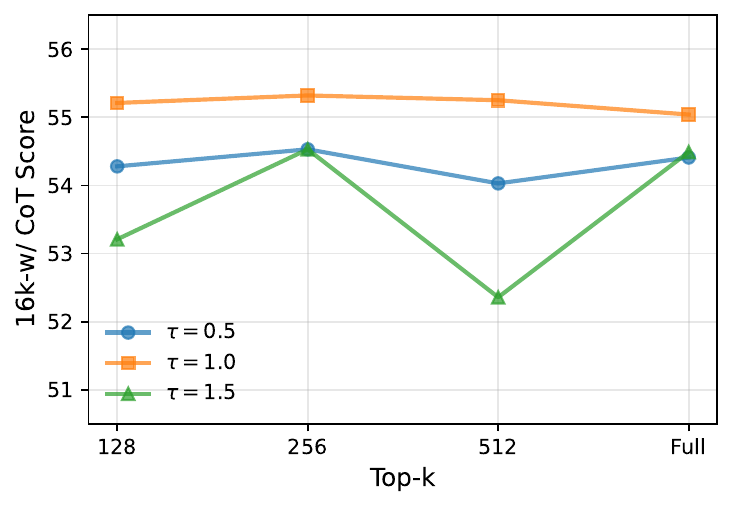}
    \caption{
    Performance of InfoKV with different $\tau$ and top-$k$. ``Full'' means that it computes entropy on the original full vocabulary.
    }

    \label{fig:ablation_topk_tau}
\end{figure}

\section{Analysis}
We conduct further studies regarding the choice of $\tau$, top-$k$ restricted entropy, and the balance between entropy and attention in this section. Furthermore, we exploit a variant of layer-wise adaptive budgets for InfoKV.

\subsection{Ablation Studies}

\paragraph{Choice of $\tau$.}
We study the effect of the bias term $\tau$ on Llama-3.1-8B-Instruct under the 40\% retaining ratio. Results on the CoT setting of LongReason are presented in Figure~\ref{fig:ablation_topk_tau}. As $\tau$ increases, the contribution of layer-wise representation distance is gradually reduced, making the entropy score for each layer rely more on predictive uncertainty from the final layer. We observe that $\tau=1$ achieves the best overall performance while also providing more stable behavior across different settings. Therefore, we adopt $\tau=1$ as the default configuration in experiments.

\paragraph{Top-$k$ Restricted Entropy.}
Moreover, we investigate the effect of top-$k$ restricted entropy by varying the value of $k$ in Figure~\ref{fig:ablation_topk_tau}. Overall, restricting entropy computation to the most probable tokens consistently improves performance compared with computing entropy over the entire vocabulary. In particular, $k=256$ achieves the best performance under different values of $\tau$. This result suggests that low-probability tokens contribute limited useful information to uncertainty estimation and may introduce noise into entropy-based token importance measurement.

\begin{figure}[t]
    \centering
    \includegraphics[width=0.48\textwidth]{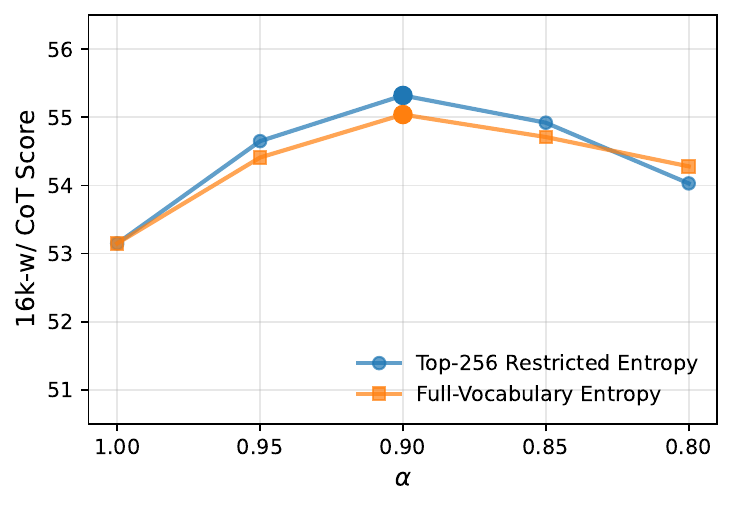}
    \caption{
    Performance of InfoKV with top-256 restricted entropy and full-vocabulary entropy across a range of $\alpha$.
    }

    \label{fig:ablation_alpha}
\end{figure}

\paragraph{Balance between Entropy and Attention.}

We further study the balance between entropy-based uncertainty and attention-based relevance by varying the coefficient $\alpha$ in Equation~\ref{eq:balanced_score}. Results are shown in Figure~\ref{fig:ablation_alpha}. When $\alpha=1$, the importance score degenerates to pure attention-based selection, which leads to inferior performance compared with using a moderate combination of entropy and attention. This observation indicates that attention alone is insufficient to fully characterize the long-range utility of tokens in KV cache.

Introducing entropy information consistently improves performance, and $\alpha=0.9$ achieves the best results under both the full-vocabulary and top-256 restricted entropy settings. However, further reducing $\alpha$ leads to performance degradation, suggesting the importance of short-range dependency from the attention perspective. Therefore, the results demonstrate that entropy and attention provide complementary signals, and a moderate integration of the two achieves the best balance of KV cache compression for long reasoning.

\subsection{Adaptive Compression}

Although the uniform strategy is effective, we observe that different Transformer layers exhibit substantially different entropy distributions. Early and middle layers generally contain richer uncertainty and broader contextual information, whereas higher layers become increasingly confident and redundant.

Motivated by this observation, we further introduce an adaptive compression strategy that dynamically allocates KV cache budgets according to layer-wise entropy statistics. Specifically, we compute the accumulated entropy score for each layer:
\begin{equation}
\bar{E}^{(l)}
=
\sum_{i=0}^{n-1}
E_i^{(l)}.
\end{equation}

The retaining budget for layer $l$ is then allocated proportionally:
\begin{equation}
k_l
=
\frac{
\bar{E}^{(l)}
}{
\sum_{m=1}^{L}
\bar{E}^{(m)}
}
\cdot
B,
\end{equation}
where $B$ denotes the total KV cache budget and $L$ is the number of Transformer layers.

Layers with larger entropy scores receive larger KV budgets, enabling the model to preserve more informative contexts in uncertainty-rich layers while aggressively compressing more redundant layers.

Table~\ref{tab:adaptive} presents the results on IFEval. Under an overall retaining ratio of 25\%, the adaptive strategy improves performance on R1-Distill-Llama-8B compared with the uniform setting. However, the gains are less consistent on R1-Distill-Qwen-7B, where adaptive allocation introduces larger performance degradation. We conjecture that excessively imbalanced layer-wise budgets may over-compress certain layers and harm the stability of long-range reasoning. Therefore, we adopt the uniform strategy as the default setting throughout the paper for better robustness and simplicity.

\begin{table}[!t]
\caption{Comparison of uniform budget and adaptive budget on IFEval.
}
\label{tab:adaptive}
\begin{center}
\begin{tabular}{@{}c@{\hspace{4pt}}c@{\hspace{4pt}}c@{\hspace{4pt}}c@{}}
\toprule
\bf Model & \bf Rate & \bf Uniform & \bf Adaptive \\
\midrule
\multirow{2}{*}{R1-Distill-Qwen-7B} & 25\% & \bf 59.70 & 55.82 \\
& 12.5\% & \bf 58.60 & 57.67 \\
\midrule
\multirow{2}{*}{R1-Distill-Llama-8B} & 25\% & 60.07 & \bf 61.18 \\
& 12.5\% & \bf 63.77 & 60.81 \\

\bottomrule
\end{tabular}
\end{center}
\end{table}

\section{Conclusion}

In this paper, we revisit KV cache compression from a forward-looking perspective and introduce forward influence to measure the effect of compressed tokens on future predictive distributions. Our analysis reveals that attention weights mainly capture short-range dependencies, whereas tokens associated with high predictive uncertainty exhibit substantially stronger influence on distant future contexts. Motivated by this observation, we propose InfoKV to combine predictive entropy, layer-wise representation evolution, and attention scores for token selection during long-context reasoning. Extensive experiments on long prefilling and long decoding benchmarks demonstrate that our information-aware KV cache compression framework consistently achieves better performance than existing attention-based compression methods across multiple models and reasoning tasks.


\section*{Limitations}

Attention weights mainly focus on how close the history contexts relate to the current query. While entropy demonstrates stronger forward influence than attention-based metrics, it remains an indirect approximation of future utility rather than an explicit optimization objective. Besides, we observe that adaptive layer-wise budget allocation improves performance for some models but can destabilize reasoning performance for others, suggesting that different architectures may exhibit distinct information distributions across layers. More robust and architecture-aware allocation strategies remain an important direction for future work.


\bibliography{custom}

\appendix

\section{Visualizations of Token Scoring}
\label{adp:case}

\begin{figure}[H]
    \centering
    \includegraphics[width=1\textwidth]{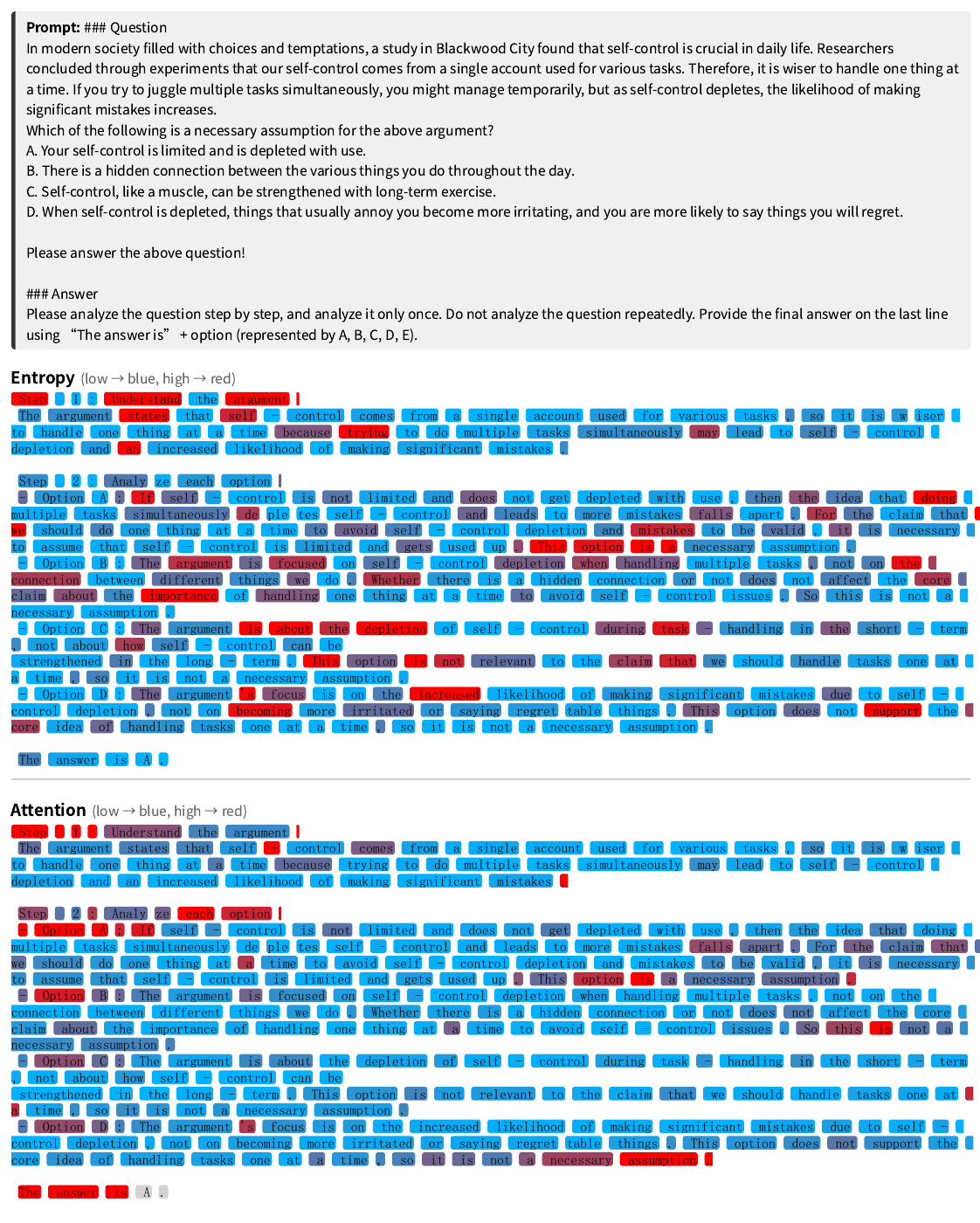}
    \caption{
    Visualizations of token scores from entropy and attention during the decoding stage of Llama-3.1-8B-Instruct on case 1. The example is a reasoning task that explicitly analyzes the question step by step.
    }
    \label{fig:case1}
\end{figure}

\begin{figure}[H]
    \centering
    \includegraphics[width=1\textwidth]{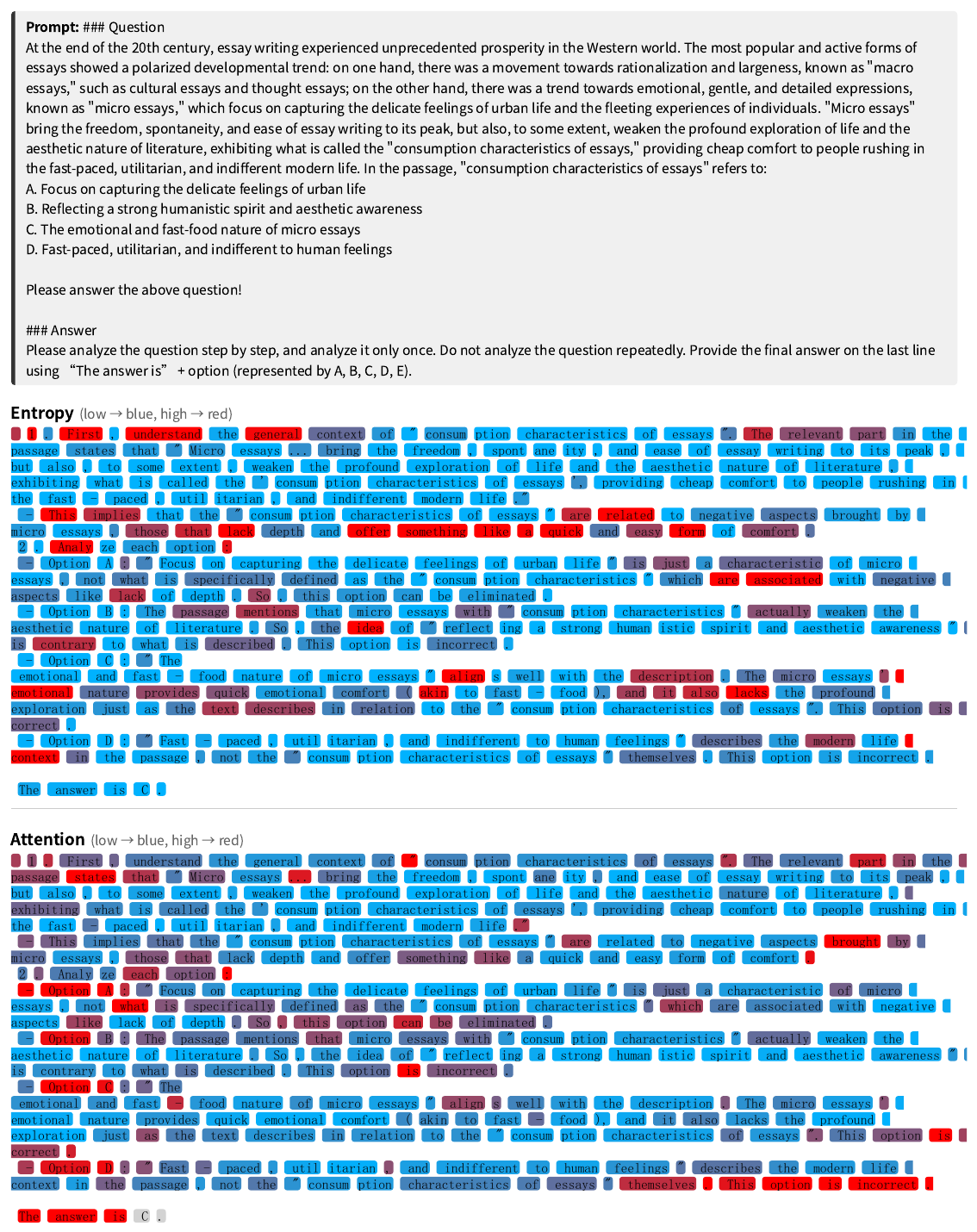}
    \caption{
    Visualizations of token scores from entropy and attention during the decoding stage of Llama-3.1-8B-Instruct on case 2.
    }
    \label{fig:case2}
\end{figure}

We provide visualizations of token scores from entropy and attention on two examples of reasoning tasks in Figures~\ref{fig:case1} and~\ref{fig:case2}.

Attention scores are obtained from ``The answer is'' in the last sequence, which is to derive the final answer. As a result, the word ``Option'' and the following ``A'', ``B'', ``C'' and ``D'' are all assigned high attention weights. It shows that attention tends to retrieve tokens that are closely relevant to the current query. In contrast, entropy measures the informativeness of the token itself, but does not depend on the query. It is observed to capture more content words that carry important information like ``argument'', ``mistakes'', ``importance'' and ``depletion''. Therefore, more information could be preserved for the future decoding process.

\section{Experiment Details}

\subsection{Long Prefill}
\label{apd:prefill_exp_setting}

For a fair comparison, all common configurations adopt the official implementations of previous KV cache compression methods, and all public hyperparameter settings are kept consistent. Specifically, the observation window size is set to 64, the pooling function adopts average pooling (\texttt{avgpool}), and the pooling kernel size is set to 9.

For InfoKV, we set the bias term $\tau$ added to $D_i^{(l)}$ to 1.0 and compute top-$k$ restricted entropy using the top 256 predicted tokens. 
To balance the weights of the entropy score and the attention score $\alpha$ in Eq.~(\ref{eq:balanced_score}) is set to 0.9. We adopt fixed prompting templates for both direct-answer and Chain-of-Thought (CoT) reasoning settings in LongReason. Same prompts are used for all compared methods to ensure that the performance differences mainly arise from KV cache compression strategies rather than prompting variations. 

\subsection{Long Decoding}
\label{apd:decode_exp_setting}

\begin{table}[!h]
\caption{Hyper-parameter settings on three long decoding tasks.}
\label{tab:hyper-parameter}
\begin{center}
\begin{tabular}{ccccc}
\toprule
\bf Model & \bf Parameter & \bf IFEval & \bf AIME 2024 & \bf LiveCodeBench \\
\midrule
\multirow{3}{*}{R1-Distill-Qwen-7B} & $\tau$ & 1.5 & 0.5 & 0.5 \\
& k & 256 & 128 & 512 \\
& $\alpha$ & 0.9 & 0.95 & 0.95 \\
\midrule
\multirow{3}{*}{R1-Distill-Llama-8B} & $\tau$ & 1 & 0.5 & 1 \\
& k & 256 & 256 & 256 \\
& $\alpha$ & 0.9 & 0.95 & 0.95 \\
\bottomrule
\end{tabular}
\end{center}
\end{table}

The settings of hyperparameters on the three long decoding benchmarks are summarized in Table~\ref{tab:hyper-parameter}. The weight of attention scores $\alpha$ is set to 0.95 for AIME 2024 and LiveCodeBench, whose samples contain a large number of mathematical notations and symbolic reasoning steps. In such cases, attention scores provide more reliable structural signals for preserving locally important contexts. For IFEval, we adopt a slightly smaller value $\alpha=0.9$ to introduce stronger entropy guidance, which better captures informative tokens relevant to instruction following and long-range code generation.


\end{document}